\documentclass[10pt,twocolumn,letterpaper]{article}

\usepackage{iccv,times,graphicx,tabulary,multirow,subfig,xspace,xcolor}
\usepackage{amsthm,amssymb,fixmath,mathtools,nicefrac}
\usepackage[pagebackref=true,breaklinks=true,letterpaper=true,colorlinks,
citecolor=citecolor,bookmarks=false]{hyperref}
\definecolor{citecolor}{RGB}{34,139,34}

\usepackage[flushleft]{threeparttable}
\usepackage{array}
\usepackage{booktabs, caption}       
\usepackage{float}

\newcommand\blfootnote[1]{%
	\begingroup
	\renewcommand\thefootnote{}\footnote{#1}%
	\addtocounter{footnote}{-1}%
	\endgroup
}

\begin{document}
	\title{Classification Calibration for Long-tail Instance Segmentation}
	\track{LVIS}
	
	
	\author{\normalsize{*Tao~Wang$^1$} \quad  
		\normalsize{*Yu~Li$^{2,1}$}   \quad 
		\normalsize{Bingyi~Kang$^{1}$} 
		\quad
		\normalsize{Junnan~Li$^{3}$}    \quad
		\normalsize{Junhao Liew$^{1}$}   \quad
		\normalsize{Sheng~Tang$^{2}$}   \quad
		\normalsize{Steven~Hoi$^{3}$} \quad
		\normalsize{Jiashi~Feng$^1$}\\
		\small{$^{1}$Department of Electrical and Computer Engineering, National University of Singapore, Singapore} \\
		\small{$^{2}$ Institute of Computing Technology, Chinese Academy of Sciences, China} \\
		\small{$^{3}$ Salesforce Research Asia, Singapore} \\
		{\small \tt twangnh@gmail.com}  \ \ 
		{\small \tt liyu@ict.ac.cn}  \ \ 
		{\small \tt bingykang@gmail.com}  \ \ 
		{\small \tt junnan.li@salesforce.com}  \ \ 
		{\small \tt liewjunhao@u.nus.edu} \ \ \\
		{\small \tt ts@ict.ac.cn}\ \
		{\small \tt shoi@salesforce.com}
		{\small\tt elefjia@nus.edu.sg}\\
	}
	
	\maketitle
	
	\begin{abstract}
		This report presents our winning solution to LVIS 2019 challenge.
		Remarkable progress has been made in object instance detection and segmentation in recent years. However, existing state-of-the-art methods are mostly evaluated with fairly balanced and class-limited benchmarks, such as Microsoft COCO dataset~\cite{lin2014coco}.  
		In this report, we investigate the performance drop phenomenon of state-of-the-art two-stage instance segmentation models when processing extreme long-tail training data based on the LVIS~\cite{gupta2019lvis} dataset, and find a major cause is the inaccurate classification of object proposals. 
		Based on this observation, we propose to calibrate the prediction of classification head to improve  recognition performance for the tail classes. 
		Without much additional cost and modification of the detection model architecture, our calibration method  improves the performance of the baseline by a large margin on the tail classes.  Importantly, after the submission, we find significant improvement can be further achieved by modifying the calibration head~\cite{wang2020devil}, codes and models are available at \url{https://github.com/twangnh/SimCal}.\blfootnote{Both * authors contributed equally to this work.}.
		
	\end{abstract}
	
	
	\section{Experimental Details}
	
	\paragraph{Dataset statistics}
	Different from~\cite{gupta2019lvis}, we divide all the 1,230 categories of the LVIS v0.5 dataset into 4 sets, which respectively contain $\textless$ 10, 10-100, 100-1,000 and $\textgreater$ 1,000 training object instances. 
	We denote them as subset (0, 10), subset [10, 100), subset [100, 1000) and subset [1000, -] for convenience of expression.
	Please see Table~\ref{data_cat_distribution} for detailed statistics.
	Beyond the test set results, we evaluate model performance based on such category split in this report, in order to see the effect of training instance number and analyze the long-tail object instance detection models. We claim that the improvement on the tail bin, i.e. subset (0, 10), of the validation set does not contribute much to the overall AP as it contains only 67 classes, though the category distribution of the test set is unknown.

	\paragraph{Training and Evaluation}
	
	Our implementation is based on the mmdetection toolkit~\cite{mmdetection}. 
	Unless otherwise stated, the models are trained on LVIS-v0.5 training set and evaluated on LVIS-v0.5 validation set for mask prediction tasks. 
	The external data used in the experiments are introduced in Sec.~\ref{external_data}. 
	All the models are trained with SGD,  0.9 momentum and 8 images per minibatch. The training schedule is 8th/11th/12th epoch updates with learning rates of 0.01/0.001/0.0001 respectively, unless otherwise stated.
	
	\begin{table}
		\small
		\centering
		\renewcommand{\tabcolsep}{1.5pt}
		\renewcommand{\arraystretch}{1.1}
		\begin{tabular}{l|cccc|r}
			\toprule
			Sets   & $(0, 10)$  & $[10, 100)$  & $[100, 1000)$  & $[1000,-]$   & total \\
			\midrule
			Train   	 & 294   & 453 & 302 & 181 & 1230 \\ 
			Train-on-val   & 67   & 298 & 284 & 181 & 830 \\
			\bottomrule
			
		\end{tabular}
		\caption{Category division based on training instance number. \emph{Train-on-val} means the subset of categories that appear in the validation set.}
		\label{data_cat_distribution}
		
	\end{table}
	
	\section{Classification Calibration}
	We first investigate the performance degradation of the baseline Mask-RCNN~\cite{he2017mask} on tail classes. Then, based on our observations for the possible causes of this phenomenon, we propose a classification calibration method for improving the model performance over tail classes. 
	
	\subsection{Missed Detection of Tail Classes}
	
	\begin{table}
		\small
		\centering
		\renewcommand{\tabcolsep}{0.5pt}
		\renewcommand{\arraystretch}{1.1}
		\begin{tabular}{l|cccc|c}
			\toprule
			Model   & AP$_{(0, 10)}$  & AP$_{[10, 100)}$  & AP$_{[100, 1000)}$  & AP$_{[1000,-]}$   & AP \\
			\midrule
			mrcnn-r50-thr   & 0.0   & 5.4 & 16.6 & 25.1 & 13.1 \\ 
			mrcnn-r50      & 0.0   & 13.3 & 21.4 & 27.0 & 18.0 \\  
			\bottomrule
		\end{tabular}
		\caption{Performance of baseline Mask-RCNN with \emph{class agnostic box and mask heads} on validation set. mrcnn-r50-thr means testing with 0.05 detection threshold and mrcnn-r50 denotes testing with 0.0 threshold.}
		\label{baseline_mrcnn_ag}
		
	\end{table}
	
	For simplicity of analysis, we train a baseline Mask R-CNN with ResNet50-FPN backbone and \emph{class agnostic box and mask heads}. 
	As shown in Table~\ref{baseline_mrcnn_ag}, the model performs poorly, especially on the tail sets (0,10] and (10, 100]. 
	Even when we lower the detection threshold to 0, which improves 5\% mAP, the mAP for the subset (0, 10] is still 0. 
	This result reveals that the Mask-RCNN model trained with normal setting is heavily biased toward the many-shot classes (i.e. those with more training instances).

	We then calculate the proposal recall of the model and compare it with that of the same model trained on COCO dataset. 
	As shown in Table~\ref{ap_ar_coco_lvis}, the same baseline model trained on LVIS suffers a drop of 8.8\% in the proposal recall compared with that on COCO, and notably, a 45.1\% drop in the overall AP. 
	This indicates the degradation of proposal classification accuracy is the major cause of final performance drop on long-tail training data.

	To verify our observation, 
	{for RPN generated proposals, we assign their ground truth class labels and evaluate the AP,  instead of using the  predicted labels.}
	As shown in Table~\ref{props_gt_label}, AP on tail classes is increased by a large margin, especially on the (0,10) and [10, 100) bins. 
	This confirms the observation that the low performance over tail classes is mainly caused by the inability of the model to recognize their correct categories from current generated proposal candidates.
	
	\begin{table}
		\small
		\centering
		\renewcommand{\tabcolsep}{2.0pt}
		\renewcommand{\arraystretch}{1.1}
		\begin{tabular}{l|cc}
			\toprule
			Dataset   & AP   &  AR$_{1k}$ \\
			\midrule
			LVIS   & 18.0   & 51.0 \\ 
			COCO   & 32.8   & 55.9 \\   
			\bottomrule
		\end{tabular}
		\caption{Comparison of baseline models trained on COCO and LVIS. Models are all evaluated with the 5k validation set. AR$_{1k}$ denotes average recall at 1000 proposals. COCO results are measured on minival set.}
		\label{ap_ar_coco_lvis}
		
	\end{table}
	
	\begin{table}
		\small
		\centering
		\renewcommand{\tabcolsep}{0.5pt}
		\renewcommand{\arraystretch}{1.1}
		\begin{tabular}{l|cccc|c}
			\toprule
			Model   & AP$_{(0, 10)}$  & AP$_{[10, 100)}$  & AP$_{[100, 1000)}$  & AP$_{[1000,-]}$   & AP \\ 
			\midrule
			mrcnn-r50   & 0.0   & 13.3 & 21.4 & 27.0 & 18.0 \\ 
			props-gt      & 39.7   & 45.1 & 31.4 & 29.3 & 36.6 \\   
			\bottomrule
		\end{tabular}
		\caption{Test with ground truth labels of proposals.}
		\label{props_gt_label}
		
	\end{table}

	\subsection{Classification Calibration with Retrained Head}
	Recently,  Kang \emph{et al.}~\cite{kang2019decoupling} reveal that for long-tail classification, representation and classifier learning should be decoupled. Inspired by this work,
	to improve the performance of the second stage classification over tail classes, we develop a strategy to retrain the classification head with data obtained by class balanced sampling and combine predictions of the new classification head with the original one.  Our approach, though simple, can effectively improve the recognition accuracy on tail classes while maintaining good performance on many-shot classes. 
	We name it \emph{classification calibration}. 

	Concretely, we sample a fixed number of classes for each step, and sample one image corresponding to each of the sampled classes. In our current implementation, 16 classes and 1 image per class are sampled. 
	The sampled images are fed to the trained model, and the obtained proposals are matched with ground truth boxes using the same IOU threshold as the original detection model training.
	Only the proposals corresponding to the sampled classes are selected, together with the ground truth boxes of these classes, for training the new head; the other proposals are ignored. 
	During training, we keep the parameters in the backbone network and RPN frozen.

	As shown in Table~\ref{calibration_methods}, with the newly trained head as the proposal classifier, AP on tail-class bins (0, 10) and [10, 100) is boosted by a large margin. 
	However, due to insufficient training on many-shot classes, AP on [100, 1000) and [1000,-] drops significantly. 
	To enjoy the advantages of both new and original heads, we have tried many different ways to combine their predictions. Refer to Table~\ref{calibration_methods} for details. 
	We find that simply concatenating the predictions of the new head on tail classes ((0, 10) and [10, 100)) with those of the original head on many-shot classes ([100, 1000) and [1000,-]) yields the best results overall.

	\subsection{Generalization to Multi-stage Cascaded Models and Large Backbones}
	To further improve the overall performance, we apply the proposed calibration method to multi-stage cascaded models with more complex architectures.
	State-of-the-art cascaded model Hybrid Task Cascade~\cite{chen2019hybrid} (HTC) is utilized here.
	We find that HTC brings a large improvement over vanilla Cascaded Mask-RCNN~\cite{cai2018cascade} on LVIS dataset. See Table~\ref{cascaded_mrcnn_htc_comparison} for details.

	All the three classification heads in the three stages of the HTC framework are retrained with our proposed sampling strategy, and we average the predictions of these three new heads during inference following the original setting.
	Then, the predictions on tail classes are concatenated with the original classification results. 
	Table~\ref{generalize_to_htc} shows the results of our calibration method applied to HTC with ResNeXt-101-64d backbone.
	The scores of categories for (0, 10) bin which are predicted by the new head are concatenated with the scores of other categories predicted by the original head.

	We think it is more reasonable to take into consideration the number of classes in each bin in long-tail detection evaluation, rather than just averaging AP of all classes.
	This is because, the number of classes in each bin may vary largely and the bins with fewer classes tend to be down-weighted in overall mAP. In this sense, the importance of solving the tail problem is not obviously and directly demonstrated by using the current evaluation metric mAP. 
	For example, the validation set of LVIS v0.5 contains only 67 classes with less than 10 training instances, while the numbers are much larger for the [10, 100) bin and [100, 1000) bin, which are 298 and 284 respectively. 
	The improvement on (0, 10) bin would be down-weighted in mAP.
	
	\begin{table}[]
		\centering
		\begin{tabular}{l|c|c|c|c}
			\toprule
			\multirow{2}{*}{Models} & \multicolumn{2}{c|}{COCO} & \multicolumn{2}{c}{LVIS} \\ \cmidrule{2-5} 
			& box         & mask        & box         & mask        \\ 
			\midrule
			cascaded-mrcnn                &      45.4       &   39.1         &      28.6       &      25.9        \\ 
			htc                &      46.9       &    40.8         &     31.3        &     29.3        \\ 
			\bottomrule
		\end{tabular}
		\caption{Comparison of cascaded Mask-RCNN and Hybrid Task Cascade (HTC) on COCO and LVIS dataset validation set. The two models use the same backbone ResNeXt-101-64x4d. They are trained with 20 epochs and learning rate decay at 16th and 19th epoch.}
		\label{cascaded_mrcnn_htc_comparison}
	\end{table}

	\begin{table}
		\small
		\renewcommand{\tabcolsep}{0.5pt}
		\renewcommand{\arraystretch}{1.1}
		\begin{tabular}{l|cccc|c}
			\toprule
			Model   & AP$_{(0, 10)}$  & AP$_{[10, 100)}$  & AP$_{[100, 1000)}$  & AP$_{[1000,-]}$   & AP \\ 
			\midrule
			mrcnn-r50      & 0.0   & 13.3 & 21.4 & 27.0 & 18.0 \\ 
			rhead-only     & 8.5   & 20.8 & 17.6 & 19.3 & 18.4 \\ 
			rhead-avg      & 8.5   & 20.9 & 19.6 & 24.6 & 20.3 \\ 
			rhead-det      & 8.6   & 22.0 & 16.7 & 25.2 & 19.8 \\ 
			rhead-cat     & \textbf{8.6}   & \textbf{22.0} & \textbf{19.6} & \textbf{26.6} & \textbf{21.1} \\ 
			rhead-cat-thr      & 8.5   & 20.8 & 20.1 & 26.7 & 20.9 \\ 
			rhead-cat-scale      & 8.5   & 21.3 & 19.9 & 26.7 & 21.0 \\  
			\bottomrule
			
		\end{tabular}
		\caption{Different ways for calibrating predictions of original classification head with newly trained head. The ways we have tried include rhead-only (using only newly trained head predictions), rhead-avg (averaging predictions of original head and new head), rhead-det (using the two heads separately for detection outputs and combining them afterward, i.e., two expert models), rhead-cat (simply concatenating tail classes predictions of new head and many-shot classes predictions of original head, with (0,10) and [10, 100) for new head and [100, 1000) [1000,-] for original head), rhead-cat-thr (filtering new head predictions with 0.05 threshold and then concatenating), and rhead-cat-scale (scaling new head predictions by ratio of average background score between new and original head predictions).}
		\label{calibration_methods}
		
	\end{table}
	
	\begin{table}
		\small
		\renewcommand{\tabcolsep}{0.5pt}
		\renewcommand{\arraystretch}{1.1}
		\begin{tabular}{l|cccc|c}
			\toprule
			Model   & AP$_{(0, 10)}$  & AP$_{[10, 100)}$  & AP$_{[100, 1000)}$  & AP$_{[1000,-]}$   & AP \\ 
			\midrule
			
			htc-x101      & 7.1   & 30.5 & 30.7 & 33.9 & 29.4 \\ 
			calibration     & \textbf{16.0}   & 30.6 & 29.8 & 33.5 & 29.8 \\  
			\midrule
			htc-x101-ms-dcn  & 5.6   & 33.0 & 33.7 & 37.0 & 31.9 \\ 
			calibration     & \textbf{12.7}   & 32.1 & 33.6 & 37.0 & 32.1 \\  
			\bottomrule
		\end{tabular}
		\caption{Results of applying our calibration to state-of-the-art multi-stage cascaded instance segmentation model Hybrid Task Cascade (HTC).}
		\label{generalize_to_htc}
		
	\end{table}

	\begin{table}
		\small
		\centering
		\renewcommand{\tabcolsep}{0.5pt}
		\renewcommand{\arraystretch}{1.1}
		\begin{tabular}{l|cccc|c}
			\toprule
			Model   & AP$_{(0, 10)}$  & AP$_{[10, 100)}$  & AP$_{[100, 1000)}$  & AP$_{[1000,-]}$   & AP \\ 
			\midrule
			mrcnn-r50      & 0.0   & 13.3 & 21.4 & 27.0 & 18.0 \\ 
			img-sample     & 7.7   & 23.2 & 21.4 & 26.2 & 22.0 \\ 
			calibration     & 8.6   & 22.0 & 19.6 & 26.6 & 21.1 \\ 
			\midrule
			htc-x101      & 5.6   & 33.0 & 33.7 & 37.0 & 31.9 \\ 
			img-sample     &  10.3  & 32.4 & 33.4 & 36.6 & 31.9 \\ 
			calibration     & 12.7   & 32.1 & 33.6 & 37.0 & 32.1 \\  
			\bottomrule
		\end{tabular}
		\caption{Comparison with image level sampling trained model on  baseline Mask R-CNN.}
		\label{compare_img_sampling}
		
	\end{table}
	
	\subsection{Comparison with Image-level Repeat Sampling}
	As shown in Table~\ref{compare_img_sampling}, we compare our classification calibration approach with image-level repeat sampling for the whole network, which is reported as the best baseline in~\cite{gupta2019lvis}.
	Although our calibration method has lower overall mAP on validation set than image-level repeat sampling on tail classes, it has higher performance on the most tail bin (0, 10). 
	When generalized to the more complex multi-stage model HTC, our method performs better.
	The performance of our calibrated model suffers less drop on many-shot classes and enjoys much improvement on tail-classes than image-level repeat sampling method.

	\section{Final Models and Test Set Submission}
	As shown in Table~\ref{final_models_ensemble_submission}, our final submitted results on the test set are from the ensemble of 4 models with different backbones.  
	However, due to time limit, we only have our best single model (31.9 AP on val) calibrated among all final models.
	We believe the final ensemble results will be stronger on tail classes if all models are calibrated.
	
	\begin{table}
		\small
		\centering
		\renewcommand{\tabcolsep}{2.0pt}
		\renewcommand{\arraystretch}{1.1}
		\begin{tabular}{l|cccc}
			\toprule
			Model   & val-set    \\ 
			\midrule
			htc\_x101\_64d\_ms\_dcn & 31.9   \\ 
			htc\_x101\_32d\_ms\_dcn   & 31.4    \\ 
			htc\_x101\_64d\_ms\_dcn\_cos    &30.7    \\  
			htc\_r101\_ms\_dcn    & 30.0   \\  
			ensemble-with-calibration    & 34.2  \\
			add-multiscale-testing    & 35.2   \\ 
			\bottomrule

		\end{tabular}
		\caption{Final models performance and ensemble results on validation set. ms denotes multi-scale training, dcn means deformable convolution and cos means cosine learning rate schedule.}
		\label{final_models_ensemble_submission}
		
	\end{table}
	
	\begin{table}
		\small
		\centering
		\renewcommand{\tabcolsep}{2.0pt}
		\renewcommand{\arraystretch}{1.1}
		\begin{tabular}{l|cccc}
			\toprule
			Model   & AP   &  AP$_{r}$ &  AP$_{c}$ &  AP$_{f}$\\ 
			\midrule
			best-baseline & 20.5   & 9.8 & 21.1 & 30.0\\ 
			w/o   & 22.89   & 5.90 & 25.65 & 35.26\\ 
			with-calibration    & 23.94   & 10.31 & 25.26 & 35.16\\  
			ensemble-with-calibration    & 26.11   & 11.94 & 27.98 & 37.05\\
			add-multiscale-testing    & 26.67   & 10.59 & 28.70 & 39.21\\
			\bottomrule

		\end{tabular}
		\caption{Comparison of baseline model without and with our proposed calibration method on LVIS \emph{test set}. Best-baseline denotes best baseline performance reported~\cite{gupta2019lvis}; w/o denotes our best single model (31.9 AP on validation set); with-calibration means adds calibration to the model; ensemble-with-calibration means using the ensemble of all models and adding calibration; add-multiscale-testing denotes adding multi-scale testing.}
		\label{htc_singlemodel_testset}
		
	\end{table}
	
	\section{External Data}
	\label{external_data}
	
	Microsoft COCO dataset~\cite{lin2014coco} (2017 version) and COCO-stuff dataset~\cite{caesar2018coco} are used as external data for our submitted results.
	All COCO, COCO-stuff, and LVIS datasets share the same training images but own different annotations (LVIS only uses part of the training images in COCO train2017).
	We only use the training set of COCO and COCO-stuff, which contains 118K images.
	COCO covers 80 thing classes, the same as COCO-stuff, but the latter also contains 91 stuff classes.
	For COCO, instance-level boxes and polygons are used to pre-train HTC models.
	We initialize our model with a model pre-trained on COCO.
	For COCO-stuff, pixel-level semantic segmentation labels are used for training the semantic head of HTC.
	Table~\ref{external_data_coco} shows the results of using COCO pre-training and semantic head.
	
	\begin{table}
		
		\small
		\centering
		\renewcommand{\tabcolsep}{0.5pt}
		\renewcommand{\arraystretch}{1.1}
		\begin{tabular}{lcc|cccc|c}
			\toprule
			Model & P & S   & AP$_{(0, 10)}$  & AP$_{[10, 100)}$  & AP$_{[100, 1000)}$  & AP$_{[1000,-]}$   & AP \\ 
			\midrule
			HTC-x50-fpn  & &     & 1.4   & 23.9 & 25.3 & 30.4 & 24.0 \\ 
			HTC-x50-fpn  & \checkmark &   & 3.9   & 25.0 & 27.0 & 31.2 & 25.3 \\ 
			HTC-x50-fpn   & \checkmark & \checkmark  & 3.7   & 27.2 & 27.4 & 31.8 & 26.4 \\ 
			\bottomrule
			
		\end{tabular}
		\caption{Effect of external data. \emph{P} stands for using COCO box and polygons for pre-training, and \emph{S} for using COCO-stuff pixel-level semantic segmentation label for semantic head of HTC.}
		\label{external_data_coco}
		
	\end{table}
	
	\section{Conclusion}
	We propose a classification calibration method for improving the performance of current state-of-the-art proposal based object instance detection and segmentation models over long-tail distribution data.
	It is able to effectively improve the classification performance over the long-tail distribution data by enhancing the proposal classification accuracy.
	Currently, our retraining strategy for proposal classification head is not optimized, which we will investigate in future works. For example, we may combine our method with image-level sampling, choose new head designs or use new head training sampling methods, trying to further boost the model performance on long-tail data distribution.
	
	{\small \bibliographystyle{ieee_fullname} \bibliography{egbib}}
	
\end{document}